\documentclass[11pt,a4paper]{article}
\usepackage[hyperref]{acl2021}
\usepackage{times}
\usepackage{latexsym}

\usepackage{todonotes}

\usepackage{amssymb}
\usepackage{lipsum}
\usepackage{tabularx}

\usepackage{arydshln}
\usepackage[inline,shortlabels]{enumitem}

\usepackage{afterpage}

\usepackage{times}
\usepackage{xspace}
\usepackage{latexsym}
\usepackage{booktabs}
\usepackage{amssymb}
\usepackage{pifont}

\usepackage{graphicx}  
\usepackage{subcaption}
\usepackage{amsfonts}
\usepackage{multicol}
\usepackage{amsmath}
\usepackage{multirow}
\usepackage{tikz}

\newcommand{\OptCls}{BERT+GPT2}

\usepackage{paralist}
\usepackage[nameinlink]{cleveref}
\creflabelformat{equation}{#1#2#3}
\crefname{equation}{eq.}{eqs.}  
\Crefname{equation}{Eq.}{Eqs.}
\crefname{figure}{Fig.}{Figs.}  
\Crefname{figure}{Fig.}{Figs.}
\DeclareRobustCommand\dashed{\tikz[baseline=-0.6ex]\draw[thick,dashed] (0,0)--(0.54,0);}

\usepackage{microtype}

\aclfinalcopy 

\newcommand*\myat{{\fontfamily{ptm}\selectfont\small @}}

\title{Discriminative and Generative Transformer-based Models For Situation Entity Classification}
\author{Mehdi Rezaee ~~~ Kasra Darvish ~~~ Gaoussou Youssouf Kebe ~~~ Francis Ferraro\\
  Department of Computer Science\\
  University of Maryland Baltimore County\\
  Baltimore, MD 21250 USA \\
{\tt \{\href{mailto:rezaee1@umbc.edu}{rezaee1},\href{mailto:kasradarvish@umbc.edu}{kasradarvish},\href{mailto:mb88814@umbc.edu}{mb88814},\href{mailto:ferraro@umbc.edu}{ferraro}\}\myat umbc.edu}
  }

\date{}

\begin{document}
\maketitle
\begin{abstract}
We re-examine the situation entity (SE) classification task with varying amounts of available training data. %
We exploit a Transformer-based variational autoencoder to encode sentences into a lower dimensional latent space, which is used to generate the text and learn a SE classifier. %
Test set and cross-genre evaluations show that when training data is plentiful, the proposed model can improve over the previous discriminative state-of-the-art models. %
Our approach performs disproportionately better with smaller amounts of training data, but when faced with extremely small sets (4 instances per label), generative RNN methods outperform transformers. Our work provides guidance for future efforts on SE and semantic prediction tasks, and low-label training regimes.
\end{abstract}

\section{Introduction} 
\label{sec:introduction}
\begin{figure}[t!]
  \centering
  \includegraphics[width=1.0\linewidth]{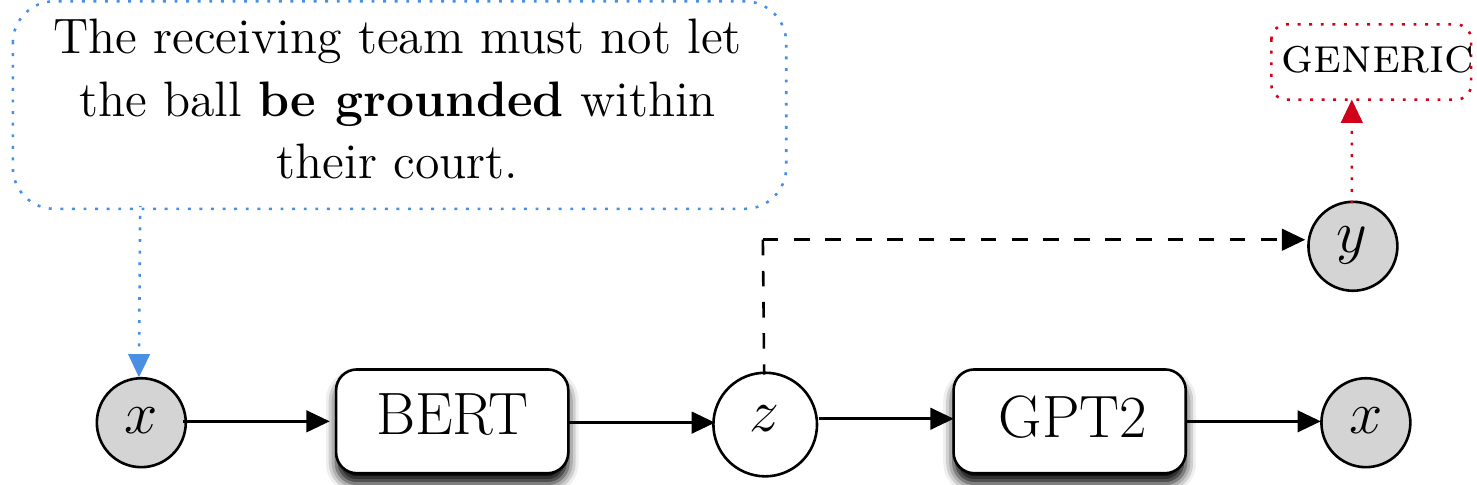}
  \caption{Model structure; $\rightarrow$ shows pre-training path for text reconstruction using unlabeled data, \dashed~ indicates label prediction during training. The
latent representation gathered from the unlabeled
data 
is
used for both text reconstruction and
SE type prediction jointly. %
}
  \label{fig:model_structure}
\end{figure}

Semantics has long recognized that a clause that references some event-like \textit{situation} may not actually be referring to any single, \textit{particular} event, but instead a general class 
of events~\citep{carlson1995generic}. For example, the sentence in \cref{fig:model_structure} provides a statement about general rules or facts (about what should not happen in a volleyball game). %
The situation that it describes is a general type of event, but it does not necessarily refer to any particular event, such as in a sentence, ``The receiving team let the ball be grounded.'' %
While linguistically intriguing, such statements can provide important background knowledge to aid various applications such as argumentation mining \citep{becker2017semantic}, discourse parsing~\citep{palmer-etal-2007-sequencing}, and characterizing genres~\citep{becker2017classifying,palmer2014genre}.

\Citet{friedrich2016situation} provide a taxonomy of, and associated dataset labeled with, 
the different types of ``situations,'' such as \textsc{event}, \textsc{state}, or a \textsc{generic} statement. %
Annotating and identifying the appropriate situation type for a clause---called \textit{situation entity (SE) typing}---is not trivial. 
While efforts to build predictive systems have steadily improved performance~\citep{friedrich2016situation,daihuang2018building}, these have only considered \textit{discriminatively} trained models. If labeled data is plentiful, discriminative models can be effective. %
Discriminative-only architectures may limit the effectiveness of these models when obtaining quality annotated data is difficult---like SE typing. %


While generatively-trained classifiers usually provide a reasonable alternative for small training sets, the performance of these generative models can degrade when trained on large training sets~\citep{ding2019latent}. %
However, the expressive capabilities of self-attention mechanism in Transformers \citep{vaswani2017attention}, make them a natural choice for both classification and language modeling tasks. %
Transformers have made it possible to effectively learn strong prior information from large-scale open-domain corpora, which is then fine-tuned on the downstream task. While often effective, a purely discriminative Transformer (BERT) has been shown to underperform the RNN-based models when training data is limited \citep{ezen2020comparison,phang2018sentence,lee2019mixout}.


Our aim in this paper is to handle the SE classification for both low and rich resource settings. We hypothesize that latent variable models and pre-trained Transformer-based models 
capture complementary information for this task. To fully make use of these two paradigms, we study a variational autoencoder model (VAE) that uses a conditional model (BERT) to encode the input sentences into a latent space, then employs a generative language model (e.g., GPT-2 or an LSTM) to regenerate the input text from that latent space \cite{li2020optimus,kingma2013auto}. %
We make the following contributions: %
\begin{enumerate*}[(1)]
\setlength{\itemsep}{0pt}
\item We show that when labeled data is plentiful, transformer methods that make local, independent predictions are able to outperform the current SOTA on SE typing that make global, joint predictions. 
\item We demonstrate that our variational approach is adaptable to lower-resource settings: when the number of samples per label is small, simpler decoders (e.g., BERT+BOW, BERT+LSTM) outperform a more complex one (i.e., BERT+GPT2).
\item We identify approaches that can nearly double performance when the number of samples per label is very small: we find generative RNN-based models outperform transformer-based models.



\end{enumerate*}
\section{Task Overview} 
\label{sec:task-overview}
We consider the task of SE type classification using
the 
publicly available MASC+Wiki dataset
\citep{friedrich2016situation}.\footnote{\url{https://github.com/BaokuiWang/context_aware_situation_entity}} %
In this task, an English clause, containing an event-like predicate, is classified into different SE types: states, events, reports, generic sentences,
generalizing sentences, questions, and imperative sentences. %
\cref{fig:model_structure} shows an example of a \textsc{generic}. %
See \Cref{Table:datasets} in the appendix for high-level statistics about the dataset, and \citet{friedrich2016situation} for an in-depth explanation of each of the SE types. %
This dataset also segments the documents into their various genres, such as ``email,'' ``fiction,'' and ``technical.''

\subsection{Related Work on SE Classification}
\label{sec:relatedWork}

Drawing inspiration from \citet{smith2003modes} and \citet{palmer-etal-2007-sequencing}, \citet{friedrich2016situation} introduced the MASC+Wiki corpus, consisting of more than $40,000$ SE-labeled clauses from $13$ different genres, and a non-neural, linear chain CRF. %
They employed a linear chain CRF based on a feature set including the POS tags and the main verb of each clause to predict the labels.
\Citet{becker2017classifying} proposed a model that combines a GRU and attention layer to capture the dependency between the tokens, labels and genres, in order to better predict the SE types. %
\Citet{daihuang2018building}, the current SOTA, used a hierarchical LSTM-based structure followed by a CRF mechanism to go beyond clause-level to achieve paragraph-wide dependencies.

While SE type prediction has been an understudied area, as \citet{friedrich2016situation} describe, one possible outcome of being able to identify the type of situation a clause is describing is an improved ability to analyze the different types of discourse~\citep{smith2003modes,palmer-etal-2007-sequencing,friedrich-palmer-2014-situation} or narration that can occur when one situation type is predominant over others. %
SE types have also been used to perform temporal analysis of situations~\citep{vempala-etal-2018-determining}, analyze participants in a reported event~\citep{sanagavarapu-etal-2017-determining}, and has helped seed deep, nuanced views of lexical semantics~\citep{govindarajan2019decomposing}.

\section{Method} 
\label{sec:method}
We study a simple yet effective encoder-decoder architecture, shown in \cref{fig:model_structure}.
For text $x$ and label $y$, we use neural variational inference (NVI) to compute a latent representation $z$ that is jointly trained to generate $x$ ($p(x | z)$) and accurately predict $y$ ($p(y | z)$). %
This yields the model $p(x, y, z) = p(x\vert z)p(y\vert z)p(z)$. NVI learns a variational distribution $q(z | x)$ to be close to the posterior $p(z | x, y)$. %



For training, we use the latent representation, computed from $q(z|x)$, for both clause reconstruction ($\mathbb{E}_{q(z|x)}\log p(x|z)$) and label prediction ($\mathbb{E}_{q(z|x)}\log p(y|z)$). %
To reduce the KL vanishing issue~\citep{bowman2016generating,shao2020controlvae}, we optimize the annealed ELBO,
\begin{equation}
\mathcal{L}  
= 
\mathbb{E}_{q(z|x)}[\log p(y|z)p(x\vert z)]
-
\beta\text{KL} [q(z|x) || p(z)],
\label{eqn:optcls-objective}
\end{equation}
where $\beta$ is the annealing coefficient fixed to $0.5$. %
To predict, we use a MAP estimation, $\mathbb{E}_{q(z|x)} \log p(y|z) \approx \log p(y\vert \mu_{z})$, where $p(y | z)$ is a single dense layer followed by softmax. 



The latent variable $z$, drawn from a neural-parametrized Gaussian distribution, captures the high-level representation of the sentence's content. %
In this setting, $q(z\vert x)$ is a multivariate Gaussian, whose mean $\mu_{z_l}$ is computed via the \textsc{cls} embedding from 
BERT, and $p(x\vert z)$ is computed by a generative text decoder seeded by $z$---for example, GPT2 or an LSTM seeded by the sampled $z$. %
Two linear functions map the latent vector to the GPT2's memory and the embedding.

The fundamental formal model we study---$p(x, y, z) = p(z) p(y | z) p(x | z)$---is a fairly straightforward latent variable model. %
While using a contextualized model to encode text and then a generative model to reconstruct it is a fairly understudied area, \citet{li2020optimus} demonstrated the effectiveness on standard classification tasks of stitching together BERT and GPT2 via a latent variable $z$, and then further pre-training the combination on English Wikipedia.\footnote{The pre-trained model can be found here: \url{https://github.com/ChunyuanLI/Optimus}} %
The core novelty of our work lies not in the precise model/inference formulation, but rather in the application and analysis of this fairly general approach to both the SE task, and the SE task at varying levels of label availability.

\section{Experiments} 
\label{sec:experiments}
\begin{table}[t]
    \centering
    \scalebox{0.85}{    
    \begin{tabular}{l|cc}
    \specialrule{.1em}{.05em}{.05em} 
    \multicolumn{1}{c|}{Model}     & F1 & \multicolumn{1}{l}{Acc} \\ \hline
    Context Aware \cite{daihuang2018building}    & 77.4               & 80.7                      \\ \cdashline{1-3}
    Discriminative & 58.3 & 69.1 \\
    Generative & 59.7 & 66.7 \\
    Latent & 60.0 & 67.0 \\   \cdashline{1-3}
    Par BERT \cite{cohan2019pretrained}& {78.3} & {81.2} \\ 
    BERT+BOW & 77.0 & 80.0 \\    
    BERT+LSTM & 77.4 & 80.1 \\
    BERT & {78.8} & {81.1} \\ 
    \OptCls{} &\textbf{79.1} & \textbf{81.9}\\
    \specialrule{.1em}{.05em}{.05em} 
    \end{tabular}
    }
    \caption{Classification performance of our methods and baselines against the current SOTA~\citet{daihuang2018building}, trained with full training data and evaluated on the test set. BERT and \OptCls{} predict each clause's SE type individually; \citet{daihuang2018building} and Par BERT predict all clauses in a paragraph jointly.}
    \label{Table:test-result}    
\end{table}
We first compare the Transformer-based models against previously proposed approaches in terms of classification metrics. We then evaluate the generative models 
with limited training data samples per SE type. We report the average accuracy over 5 runs and compare the accuracy of all models.\footnote{See \Cref{sec:additional-impl-models} for more details on our implementation, models, and baselines.} %
%


\paragraph{Models}
The central instantiation of the \cref{sec:method} method uses BERT to encode and GPT-2 to regenerate the text. %
We study the effectiveness of the generative model, by presenting two variants of our model with different decoders. Specifically, we redefine $p(x | z)$ in two, simpler, light-weight ways:
\begin{enumerate*}[label=(\arabic*)]
\setlength{\itemsep}{0mm}
\item \textbf{BERT+BOW}, where we use the simple bag-of-words method for the reconstruction part.
\item \textbf{BERT+LSTM}, \end{enumerate*}%
where we use a single LSTM decoder layer as a much simpler autoregressive alternative to GPT2. %
In BERT+BOW, $z$ is passed to a single, linear softmax, while BERT+LSTM uses $z$ to help compute each token's hidden state.

\paragraph{Baselines}
We compare our variational method from \cref{sec:method} with the following approaches:
\begin{enumerate*}[label=(\arabic*)]
\setlength{\itemsep}{0mm}
\item The current SOTA \textbf{Context Aware}~\citep{daihuang2018building} for SE prediction:
A paragraph is fed to a word-level Bi-LSTM with 300 hidden units. 
Max pooling over the Bi-LSTM hidden vectors extracts clause representations and another Bi-LSTM uses the clause representations to predict SE types.

\item \textbf{BERT}: We perform clause-level classification by adding a fully connected layer followed by a softmax classifier on top of  the pretrained uncased BERT-Base model. %

\item \textbf{Par BERT}~\citep{cohan2019pretrained}:
Instead of processing clauses of a paragraph one-by-one, the entire paragraph is provided to BERT. Clauses are separated by \texttt{[SEP]}, and an MLP uses each of the \texttt{[SEP]} embeddings to predict a clause's SE type.

\end{enumerate*}
These baselines allow use to compare to the SOTA, basic transformer methods, and transformer methods specifically designed for longer text sequences (such as paragraphs). %

However, based on recent work in low-resource text classification~\citep{ding2019latent}, we additionally consider three new baselines: %
\begin{enumerate*}[resume,label=(\arabic*)]
\setlength{\itemsep}{0mm}
\item \textbf{Discriminative Model}~\citep{yogatama2017generative}: A one-layer LSTM model
encodes the sentences and a softmax layer over the average of hidden vectors predicts the labels.
\item \textbf{Generative Model}~\citep{yogatama2017generative}:
Each label has an embedding. The tokens are fed to a one-layer LSTM model, and concatenated label embeddings and hidden vectors are used to reconstruct the tokens. %
\item \textbf{Latent Model}~\citep{ding2019latent}:
This LSTM-based method computes $p(y, x) = \sum_c p(y, c, x)$ via a discrete latent variable $c$. We let $c$ be a 30 dimensional variable. 
\end{enumerate*}
\begin{table}[!t]
    \centering
    \resizebox{.98\columnwidth}{!}{        
    \begin{tabular}{l|ccccc}
    \specialrule{.1em}{.05em}{.05em} 
    Genre     & Context Aware & BERT & Par BERT & \OptCls{}    &Humans \\ \hline
    blog      & 70.3       &{72.03}  & \textbf{74.14} & 72.37  & 72.9 \\
    email     & 71.5       &{73.84}  & \textbf{75.88} & 74.53  & 67.0 \\
    essays    & 64.1       &{66.99}  & \textbf{67.49} & 66.15  & 64.6 \\
    ficlets   & 68.8       &\textbf{75.14}  & 73.11 & 73.86  & 81.7 \\
    fiction   & 72.1       &{77.52}  & 75.42 & \textbf{78.99}  & 76.7 \\ 
    gov-docs  & 68.9       &\textbf{72.52}  & 72.31 & 71.08  & 72.6 \\
    jokes     & 75.0       &\textbf{77.11}  & 74.46 & 76.73  & 82.0 \\
    journal   & 66.4       &{68.75}  & 68.81 & \textbf{71.96}  & 63.7 \\
    letters   & 71.2       &{72.01}  & \textbf{75.64} & 71.93  & 68.0 \\
    news      & 72.7       &\textbf{75.20}  & 74.58 & 73.11  & 78.6 \\
    technical & 60.5       &{61.72}  & 53.38 & \textbf{62.72}  & 54.7 \\
    travel    & 53.6       &\textbf{68.54}  & 54.74 & 58.18  & 48.9 \\
    wiki      & 60.6       &{63.04}  & 64.58 & \textbf{67.86}  & 69.2 \\
    \specialrule{.1em}{.05em}{.05em} 
    \end{tabular}
    }
    \caption{Cross-genre F1 Classification Results. Models are trained on all other genres (e.g., not-blog) and then evaluated on the target genre (blog). All Transformer methods surpass the current SOTA.}
    \label{Table:cross-genre}
\end{table}
\subsection{Full Training}



The comparison of all of the baselines and the proposed models are shown in \Cref{Table:test-result}. Most of the Transformer-based models outperform the previous state-of-the-art approach from \citet{daihuang2018building}. %
We want to draw particular attention to the fact that \citet{daihuang2018building}'s best model predicted clause SE types jointly for every clause in a paragraph, while BERT and \OptCls{} predictions are made independently for each clause. That is, the previous SOTA method relied on global context, whereas the Transformer methods are able to do localized prediction without the explicit global context. In addition, when we made joint predictions for all clauses in a paragraph (Par BERT), performance slightly decreased. %
These results suggest that Transformers are able to implicitly capture enough background lexical knowledge to be performant when presented with limited/reduced context, though the lack of improvement of Par BERT over BERT or \OptCls{} suggests taking advantage of broader context remains a challenge. %

We carried out cross-genre classification experiments on the training dataset (\Cref{Table:cross-genre}). %
We only compare Context Aware, BERT, Par BERT, and \OptCls{}, as those had the highest accuracy results in \Cref{Table:test-result}; we include human scores (from \citet{daihuang2018building}) for context. %
For all genres, BERT and \OptCls{} outperform the Context Aware model. For more than half of the genres, each of these two models are performing better than the human predictions. 
These results reaffirm the notion that while transformer-based approaches can be very beneficial, there exist sizable differences in performance across domains. Fully understanding these domain-differences is out-of-scope for this work and is an area for future study.

\subsection{Medium-data and Small-data Training}
\label{sec:expt:small-data}

As the expertise and overall cost for quality annotation is high, we take random subsets of the training data, having 64, 100, 400, 600, and 1000 instances (clauses) per label; while sizable, 1000 is an order of magnitude less than the full set.\footnote{The current SOTA model is paragraph-level, which reduces to an LSTM model for the clause-level task.} 

\begin{figure}[t!]
  \centering
  \includegraphics[width=1.0\linewidth]{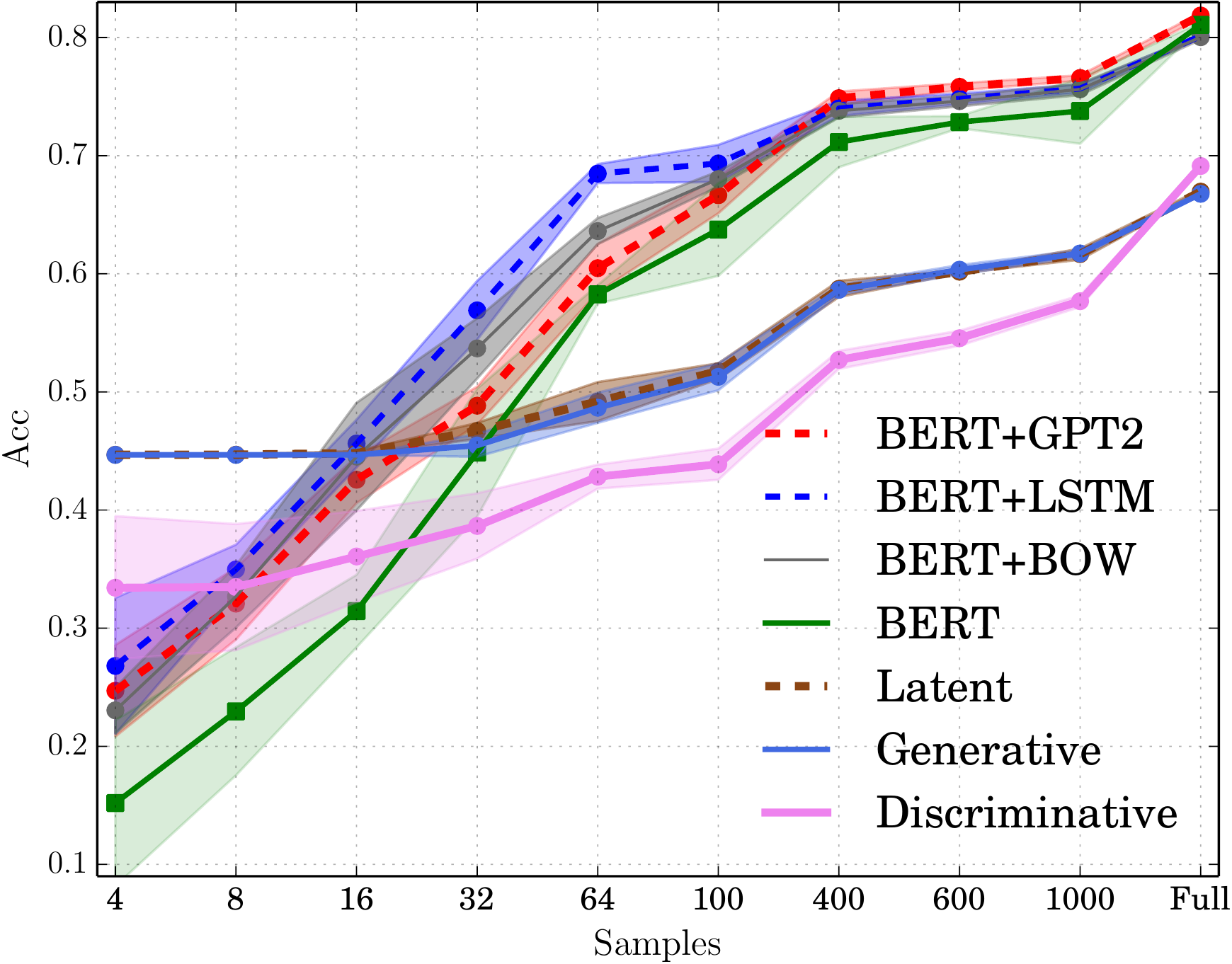}
  \caption{Performance, based on the number of instances per label during training. 
  }
  \label{fig:small-data-results}
\end{figure}

In \cref{fig:small-data-results}, with at least 64 examples per label, we observe a large gap between the Transformer-based and RNN-based models (Lat, Gen, Disc). With more than 100 training instances, the transformer methods are fairly close, though at 64 and 100 they begin to separate, with BERT underperforming the variational methods, and BERT+LSTM surpassing all others. These gaps show the effectiveness of the pre-trained models, and the effectiveness of the variational training at lower levels of supervision. 


\subsection{Extreme Low-Label Learning}
\label{sec:expt:extreme-data}

As the number of instances decreases, the accuracy of BERT becomes disproportionately lower than the other transformer methods. 
With 16-100 samples/label, BERT rapidly declines, yet the variational transformer methods provide some mitigation, e.g., at 32 samples \OptCls{} is 10\% higher than BERT. %
At even lower levels, the RNN-based Latent and Generative methods outperform the transformer methods. %
While variationally-trained transformer methods can be effective in low-resource setting, RNN methods may be better in extremely low label settings.


\section{Conclusion} 
We investigated the performance of discriminative and generative transformer models on the SE classification task, reporting new SOTA results. %
We showed that generative language modeling can be leveraged via latent variable learning into large improvements in the low-resource setting. 
Our work provides guidance and the foundation for future SE and low-label classification work.
\label{sec:conclusion}
\clearpage
\bibliographystyle{acl_natbib}
\bibliography{acl2021}

\begin{thebibliography}{24}
\expandafter\ifx\csname natexlab\endcsname\relax\def\natexlab#1{#1}\fi

\bibitem[{Becker et~al.(2017{\natexlab{a}})Becker, Palmer, and
  Frank}]{becker2017semantic}
Maria Becker, Alexis Palmer, and Anette Frank. 2017{\natexlab{a}}.
\newblock Semantic clause types and modality as features for argument analysis
  1.
\newblock \emph{Argument \& Computation}, 8(2):95--112.

\bibitem[{Becker et~al.(2017{\natexlab{b}})Becker, Staniek, Nastase, Palmer,
  and Frank}]{becker2017classifying}
Maria Becker, Michael Staniek, Vivi Nastase, Alexis Palmer, and Anette Frank.
  2017{\natexlab{b}}.
\newblock Classifying semantic clause types: Modeling context and genre
  characteristics with recurrent neural networks and attention.
\newblock In \emph{Proceedings of the 6th Joint Conference on Lexical and
  Computational Semantics (* SEM 2017)}, pages 230--240.

\bibitem[{Bowman et~al.(2016)Bowman, Vilnis, Vinyals, Dai, Jozefowicz, and
  Bengio}]{bowman2016generating}
Samuel Bowman, Luke Vilnis, Oriol Vinyals, Andrew Dai, Rafal Jozefowicz, and
  Samy Bengio. 2016.
\newblock Generating sentences from a continuous space.
\newblock In \emph{Proceedings of The 20th SIGNLL Conference on Computational
  Natural Language Learning}, pages 10--21.

\bibitem[{Carlson and Pelletier(1995)}]{carlson1995generic}
Gregory~N Carlson and Francis~Jeffry Pelletier. 1995.
\newblock \emph{The Generic Book}.
\newblock University of Chicago Press.

\bibitem[{Cohan et~al.(2019)Cohan, Beltagy, King, Dalvi, and
  Weld}]{cohan2019pretrained}
Arman Cohan, Iz~Beltagy, Daniel King, Bhavana Dalvi, and Daniel~S Weld. 2019.
\newblock Pretrained language models for sequential sentence classification.
\newblock In \emph{Proceedings of the 2019 Conference on Empirical Methods in
  Natural Language Processing and the 9th International Joint Conference on
  Natural Language Processing (EMNLP-IJCNLP)}, pages 3684--3690.

\bibitem[{Dai and Huang(2018)}]{daihuang2018building}
Zeyu Dai and Ruihong Huang. 2018.
\newblock \href {https://doi.org/10.18653/v1/D18-1368} {Building context-aware
  clause representations for situation entity type classification}.
\newblock In \emph{Proceedings of the 2018 Conference on Empirical Methods in
  Natural Language Processing}, pages 3305--3315, Brussels, Belgium.
  Association for Computational Linguistics.

\bibitem[{Ding and Gimpel(2019)}]{ding2019latent}
Xiaoan Ding and Kevin Gimpel. 2019.
\newblock Latent-variable generative models for data-efficient text
  classification.
\newblock In \emph{Proceedings of the 2019 Conference on Empirical Methods in
  Natural Language Processing and the 9th International Joint Conference on
  Natural Language Processing (EMNLP-IJCNLP)}, pages 507--517.

\bibitem[{Ezen-Can(2020)}]{ezen2020comparison}
Aysu Ezen-Can. 2020.
\newblock A comparison of lstm and bert for small corpus.
\newblock \emph{arXiv preprint arXiv:2009.05451}.

\bibitem[{Friedrich and Palmer(2014)}]{friedrich-palmer-2014-situation}
Annemarie Friedrich and Alexis Palmer. 2014.
\newblock \href {https://doi.org/10.3115/v1/W14-4921} {Situation entity
  annotation}.
\newblock In \emph{Proceedings of {LAW} {VIII} - The 8th Linguistic Annotation
  Workshop}, pages 149--158, Dublin, Ireland. Association for Computational
  Linguistics and Dublin City University.

\bibitem[{Friedrich et~al.(2016)Friedrich, Palmer, and
  Pinkal}]{friedrich2016situation}
Annemarie Friedrich, Alexis Palmer, and Manfred Pinkal. 2016.
\newblock Situation entity types: automatic classification of clause-level
  aspect.
\newblock In \emph{Proceedings of the 54th Annual Meeting of the Association
  for Computational Linguistics (Volume 1: Long Papers)}, pages 1757--1768.

\bibitem[{Govindarajan et~al.(2019)Govindarajan, Durme, and
  White}]{govindarajan2019decomposing}
Venkata Govindarajan, Benjamin~Van Durme, and Aaron~Steven White. 2019.
\newblock Decomposing generalization: Models of generic, habitual, and episodic
  statements.
\newblock \emph{Transactions of the Association for Computational Linguistics},
  7:501--517.

\bibitem[{Ide et~al.(2008)Ide, Baker, Fellbaum, Fillmore, and
  Passonneau}]{ide2008masc}
Nancy Ide, Collin Baker, Christiane Fellbaum, Charles Fillmore, and
  Rebecca~Jane Passonneau. 2008.
\newblock Masc: The manually annotated sub-corpus of american english.
\newblock In \emph{6th International Conference on Language Resources and
  Evaluation, LREC 2008}, pages 2455--2460. European Language Resources
  Association (ELRA).

\bibitem[{Kingma and Welling(2013)}]{kingma2013auto}
Diederik~P Kingma and Max Welling. 2013.
\newblock Auto-encoding variational bayes.
\newblock \emph{arXiv preprint arXiv:1312.6114}.

\bibitem[{Lee et~al.(2019)Lee, Cho, and Kang}]{lee2019mixout}
Cheolhyoung Lee, Kyunghyun Cho, and Wanmo Kang. 2019.
\newblock Mixout: Effective regularization to finetune large-scale pretrained
  language models.
\newblock \emph{arXiv preprint arXiv:1909.11299}.

\bibitem[{Li et~al.(2020)Li, Gao, Li, Peng, Li, Zhang, and Gao}]{li2020optimus}
Chunyuan Li, Xiang Gao, Yuan Li, Baolin Peng, Xiujun Li, Yizhe Zhang, and
  Jianfeng Gao. 2020.
\newblock \href {https://doi.org/10.18653/v1/2020.emnlp-main.378} {Optimus:
  Organizing sentences via pre-trained modeling of a latent space}.
\newblock In \emph{Proceedings of the 2020 Conference on Empirical Methods in
  Natural Language Processing (EMNLP)}, pages 4678--4699, Online. Association
  for Computational Linguistics.

\bibitem[{Palmer and Friedrich(2014)}]{palmer2014genre}
Alexis Palmer and Annemarie Friedrich. 2014.
\newblock Genre distinctions and discourse modes: Text types differ in their
  situation type distributions.
\newblock In \emph{Workshop on Frontiers and Connections between Argumentation
  Theory and NLP}.

\bibitem[{Palmer et~al.(2007)Palmer, Ponvert, Baldridge, and
  Smith}]{palmer-etal-2007-sequencing}
Alexis Palmer, Elias Ponvert, Jason Baldridge, and Carlota Smith. 2007.
\newblock \href {https://www.aclweb.org/anthology/P07-1113} {A sequencing model
  for situation entity classification}.
\newblock In \emph{Proceedings of the 45th Annual Meeting of the Association of
  Computational Linguistics}, pages 896--903, Prague, Czech Republic.
  Association for Computational Linguistics.

\bibitem[{Phang et~al.(2018)Phang, F{\'e}vry, and Bowman}]{phang2018sentence}
Jason Phang, Thibault F{\'e}vry, and Samuel~R Bowman. 2018.
\newblock Sentence encoders on stilts: Supplementary training on intermediate
  labeled-data tasks.
\newblock \emph{arXiv preprint arXiv:1811.01088}.

\bibitem[{Sanagavarapu et~al.(2017)Sanagavarapu, Vempala, and
  Blanco}]{sanagavarapu-etal-2017-determining}
Krishna~Chaitanya Sanagavarapu, Alakananda Vempala, and Eduardo Blanco. 2017.
\newblock Determining whether and when people participate in the events they
  tweet about.
\newblock In \emph{Proceedings of the 55th Annual Meeting of the Association
  for Computational Linguistics (Volume 2: Short Papers)}.

\bibitem[{Shao et~al.(2020)Shao, Yao, Sun, Zhang, Liu, Liu, Wang, and
  Abdelzaher}]{shao2020controlvae}
Huajie Shao, Shuochao Yao, Dachun Sun, Aston Zhang, Shengzhong Liu, Dongxin
  Liu, Jun Wang, and Tarek Abdelzaher. 2020.
\newblock Controlvae: Controllable variational autoencoder.
\newblock In \emph{International Conference on Machine Learning}, pages
  8655--8664. PMLR.

\bibitem[{Smith(2003)}]{smith2003modes}
Carlota~S Smith. 2003.
\newblock \emph{Modes of discourse: The local structure of texts}, volume 103.
\newblock Cambridge University Press.

\bibitem[{Vaswani et~al.(2017)Vaswani, Shazeer, Parmar, Uszkoreit, Jones,
  Gomez, Kaiser, and Polosukhin}]{vaswani2017attention}
Ashish Vaswani, Noam Shazeer, Niki Parmar, Jakob Uszkoreit, Llion Jones,
  Aidan~N Gomez, {\L}ukasz Kaiser, and Illia Polosukhin. 2017.
\newblock Attention is all you need.
\newblock In \emph{Advances in neural information processing systems}, pages
  5998--6008.

\bibitem[{Vempala et~al.(2018)Vempala, Blanco, and
  Palmer}]{vempala-etal-2018-determining}
Alakananda Vempala, Eduardo Blanco, and Alexis Palmer. 2018.
\newblock Determining event durations: Models and error analysis.
\newblock In \emph{Proceedings of the 2018 Conference of the North {A}merican
  Chapter of the Association for Computational Linguistics: Human Language
  Technologies, Volume 2 (Short Papers)}.

\bibitem[{Yogatama et~al.(2017)Yogatama, Dyer, Ling, and
  Blunsom}]{yogatama2017generative}
Dani Yogatama, Chris Dyer, Wang Ling, and Phil Blunsom. 2017.
\newblock Generative and discriminative text classification with recurrent
  neural networks.
\newblock \emph{arXiv preprint arXiv:1703.01898}.

\end{thebibliography}
\clearpage

\date{}


\appendix
\title{Supplementary Material}
\maketitle

\section{Dataset}

The MASC+Wiki corpus consists of more than 40,000 sentences with their corresponding SE types (labels) from Wikipedia and MASC~\citep{ide2008masc}. There are $26,283$ training data , $6,571$ validation data, and $7,937$ test data. The details are shown in \Cref{Table:datasets}.
\begin{table}[t!]
\centering
\scalebox{0.9}{
\begin{tabular}{l|rr|r}
\specialrule{.1em}{.05em}{.05em} 
SE type      & \multicolumn{1}{c}{MASC} & \multicolumn{1}{c|}{Wiki} & \multicolumn{1}{c}{Count} \\ \hline
STATE        & 49.8\%                   & 24.3\%                    & 18337                      \\
EVENT        & 24.3\%                   & 18.9\%                    & 9688                       \\
REPORT       & 4.8\%                    & 0.9\%                     & 1617                       \\
GENERIC      & 7.3\%                    & 49.7\%                    & 7582                       \\
GENERALIZING & 3.8\%                    & 2.5\%                     & 1466                       \\
QUESTION     & 3.3\%                    & 0.1\%                     & 1056                       \\
IMPERATIVE   & 3.2\%                    & 0.2\%                     & 1046                        \\
\specialrule{.1em}{.05em}{.05em} 
\end{tabular}
}
\caption{Dataset Statistics, as reported by \citet{friedrich2016situation}. The Count column shows the number of clauses per SE type.}
\label{Table:datasets}
\end{table}

\section{Additional Implementation and Model Details}
\label{sec:additional-impl-models}

\subsection{Implementation Details}

All models were trained and tested on a single RTX-based GPU. No models required more than 11GB of memory and 5 hours of runtime. %
Models were trained until convergence, as determined by performance on the validation dataset. %

\subsection{BERT and GPT2 Structures}
We use the basic (smaller) BERT model, with 12 layers, 12 self-attention heads, a hidden size of 768 and a total of 110 million parameters. %

The GPT2 layer has $L$ layers and the embedding dimension is $H$.
One linear layer maps the latent $z \in \mathbb{R}^P$ layer to $h_{\text{Mem}} = W_m z$ and the other one $h'_{\text{Emb}}=
h_{\text{Emb}} + W_{D}z
$, where $W_m \in \mathbb{R}^{LH\times P}$ projects the latent $z$ to the $L$ layers and 
$W_{D} \in \mathbb{R}^{H\times P}$ projects the latent to the embedding space.
 Our latent dimension in these experiments is $P=30$, $L=12$ and $H=768$.

\subsection{Comparison of Generative and Discriminative Classifiers}
\label{GenLatClassifiers}
The generative and latent baselines maximize the joint probability of tokens and labels.
In the discriminative classifier the conditional probability of labels given documents is maximized $\sum_{\langle x,y \rangle \in \mathcal{D}} \log p(y \vert x)$ where $x$ is encoded using an LSTM. We mention the generative classifier and the latent-variable generative model and how they differ from each other and from the discriminative classifier in following subsections.

\subsubsection{{Class-based Language Model} (Gen):} 
\label{appendix:class-based}
The objective of a generative classifier is to maximize the joint probability $\sum_{\langle x,y \rangle \in \mathcal{D}} \log p(x,y)$ where 
$x$ and $y$ represent a document of length $T$ and its label, respectively.
The factorization of the joint probability is given in equation~\ref{eq:joint-factor}.

            \begin{equation}\label{eq:joint-factor}
                p(x, y) = p(x \mid y) p(y),
            \end{equation}                
            where
            \begin{equation}
                \log p(x \mid y) = \sum_{t=1}^{T} \log p(x_t \mid x_{<t}, y)
            \end{equation} 
The prediction of the next word $x_{t+1}$ is done by concatenating the LSTM hidden states and the label embeddings and feeding them into a softmax layer.
For prediction at inference, the discriminative classifier maximizes $p(y \vert x)$ with respect to $y$ and the generative classifier maximizes
$p(x \vert y)p(y)$.

\subsubsection{Latent-Variable Generative Model (Lat):}    

Incorporating discrete latent variables into the standard generative classifier can be formulated as shown in equation~\ref{eq:latent-joint-factor},
\begin{equation}\label{eq:latent-joint-factor}
                p(x,y,c) = p_{\Theta}(x \mid c, y) p_{\Phi}(c) p_{\Psi}(y),
\end{equation}                
where $\Theta$ and $\Phi$ are the set of parameters of the language
model and the set of parameters for the prior distribution of the latent variable, respectively. Same as the generative classifier, $p_{\Psi}(y)$ is obtained from the empirical label distribution.
The prior distribution of the latent variable is parameterized in equation~\ref{eq:prior-latent}:
            \begin{align}\label{eq:prior-latent}
                p_{\Phi}(c) \propto \exp\{w_c ^\top v_c + b_c\}.
            \end{align}        
Same as the generative classifier, the prediction is done by an LSTM and a softmax layer.
            \begin{align}
                p_{\Theta}(x_t \mid x_{<t}, c, y) \propto \exp\{u^\top ([h_t; v_y; v_c]) + b\}
            \end{align}        
Here, $v_y$ and $v_c$ are the embeddings for the label and the latent variable, and $[.;.]$ indicates vertical concatenation.
The hidden representation, label, and latent embeddings are concatenated for the text reconstruction.
The training objective of the latent-variable generative model is to maximize the log marginal likelihood which is shown in equation~\ref{eq:train-objective-marginal}.

            \begin{align}\label{eq:train-objective-marginal}
                \max_{\Theta, \Phi, V_\mathcal{C}, V_\mathcal{Y}}  \sum_{\langle x,y \rangle \in \mathcal{D}} \log \sum_{c \in \mathcal{C}} p(x \mid c, y) p(c) p(y)
            \end{align}   
 All the embeddings and hidden representations are 100-dimensional in these baselines. %
 The latent model requires the expensive and time consuming full marginalization over the latent variables $c$, which is a major obstacle for using this model with Transformers.            


\begin{figure*}
  \centering
  \includegraphics[width=0.9\linewidth]{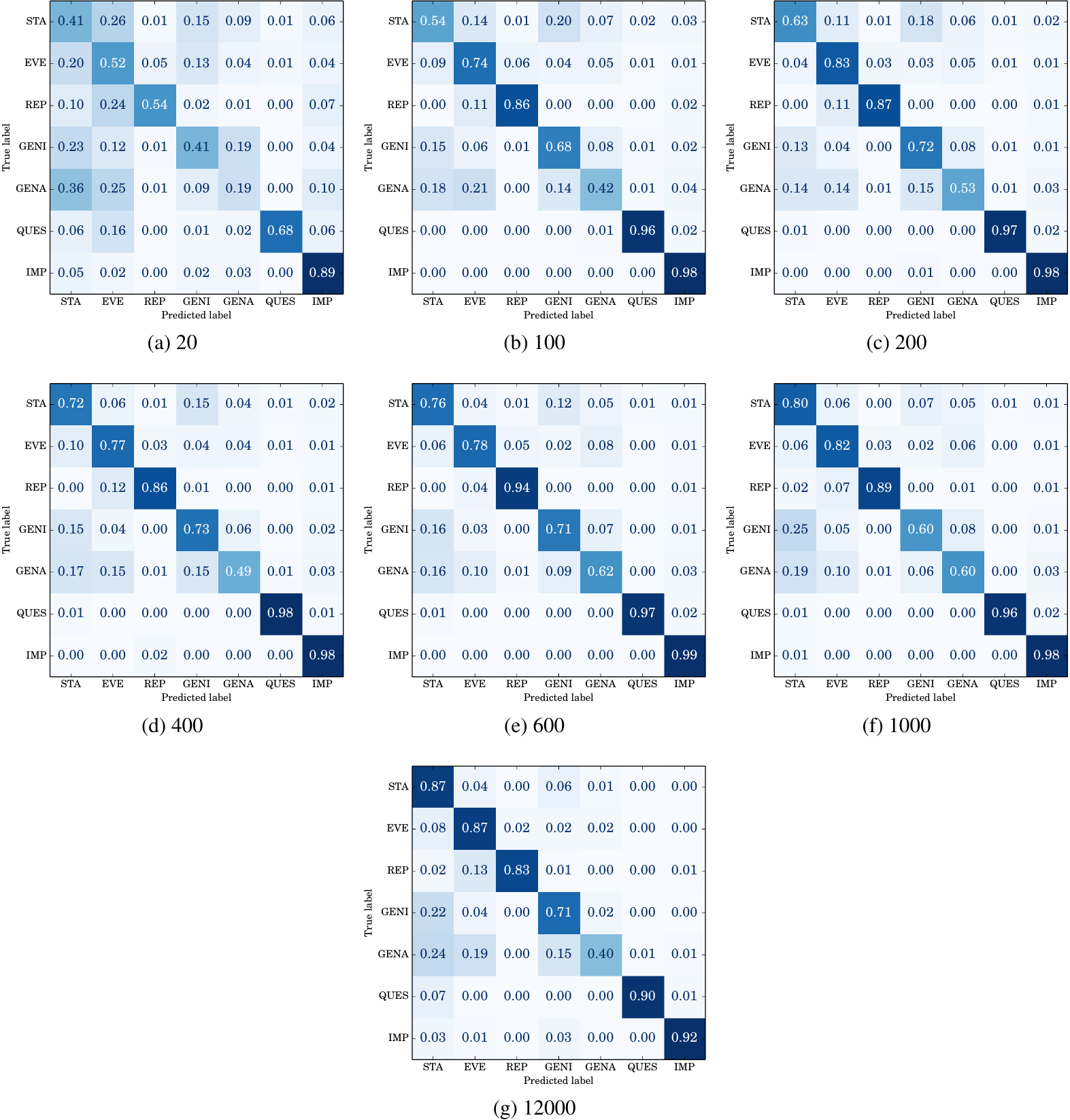}
  \caption{Confusion matrices for \OptCls{}, as we vary the number of samples per label. Notice a diagonal pattern emerges relatively quickly, though there are large confusions, especially between STATIVE and GENERIC SE labels.}
  \label{fig:tsne_train_20-confusion-matrix}
\end{figure*}

\section{Additional Insights}
\label{apx:tsne}

In this section we provide some additional insights to supplement the core results reported in the main paper.

In \cref{fig:tsne_train_20-confusion-matrix}, we show confusion matrices for \OptCls{}, as we vary the number of samples per label. Notice a diagonal pattern emerges relatively quickly, though there are large confusions, especially between STATIVE and GENERIC SE labels.

\begin{figure*}
  \centering
  \includegraphics[width=0.9\linewidth]{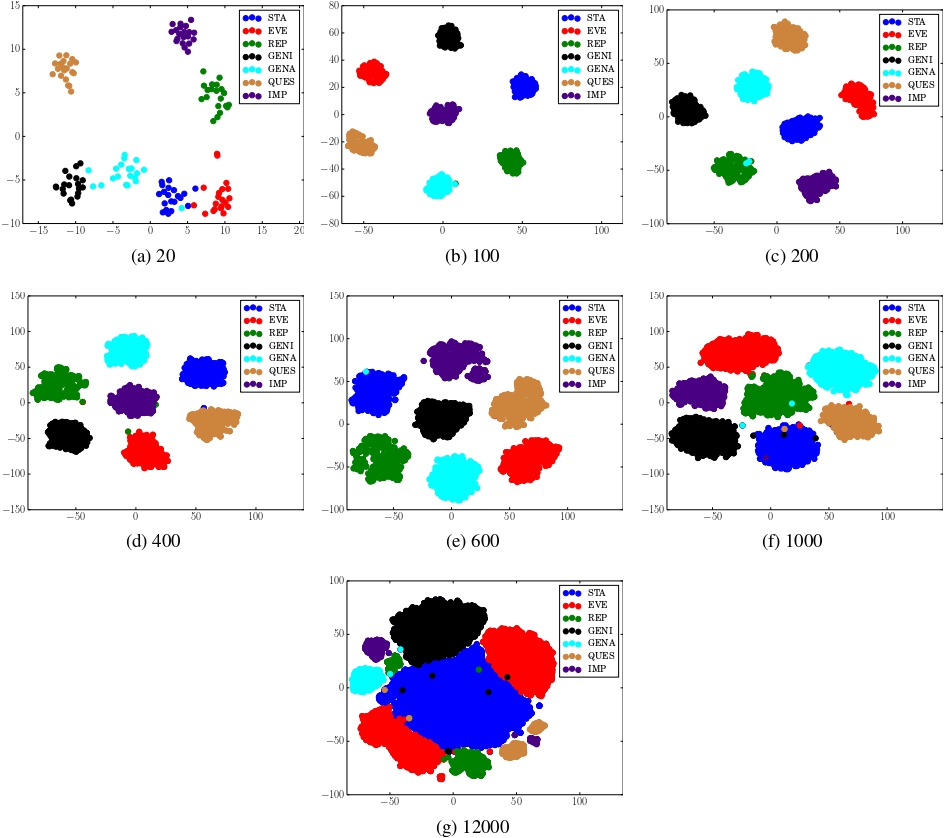}
  \caption{A series of t-SNE representations from training set clauses.}
  \label{fig:tsne_train_20-tsne}
\end{figure*}

In the paper we mentioned that the MAP approximation we use when making (and learning to make) the predictions has a nice, qualitatively observed side-effect: that the training results in latent variables $z$ that can be nicely clustered. This can be observed in \cref{fig:tsne_train_20-tsne}, where we show t-SNE plots for the mean $\mu_z$ from the training set, as we vary the amount of supervision we have per label. 

\end{document}